\documentclass[10pt,twocolumn,letterpaper]{article}

\usepackage{cvpr}
\usepackage{times}
\usepackage{epsfig}
\usepackage{graphicx}
\usepackage{amsmath}
\usepackage{amssymb}

\usepackage{times}
\usepackage{latexsym}
\usepackage{color}
\usepackage{amsmath}
\usepackage{multirow}
\usepackage{makecell}
\usepackage{graphicx}
\usepackage{multirow}
\usepackage{algorithm}
\usepackage{algorithmic}
\usepackage{amssymb}
\usepackage{amsmath}
\usepackage{dsfont}
\usepackage{float} 
\usepackage{subfigure}

\usepackage{array}
\newcolumntype{P}[1]{>{\centering\arraybackslash}p{#1}}
\usepackage{bbm}
\usepackage{siunitx}
\usepackage{caption}
\usepackage{setspace} 
\usepackage{bbding}
\usepackage{color}
\usepackage{amsmath, bm}
\usepackage{booktabs}

\usepackage{array}
\usepackage{color}
\usepackage{subfiles}
\usepackage{setspace}
\usepackage{soul} 
\usepackage{multirow}
\usepackage{caption}

\usepackage{pifont}
\usepackage{xspace}

\usepackage[pagebackref=true,breaklinks=true,letterpaper=true,colorlinks,bookmarks=false]{hyperref}

 \cvprfinalcopy 

\def\cvprPaperID{****} 

\ifcvprfinal\fi
\begin{document}

\title{Deep Reason: A Strong Baseline for Real-World Visual Reasoning}

\author{Chenfei Wu\\
{\tt\small wuchenfei@bupt.edu.cn}
\and
Yanzhao Zhou\\
{\tt\small zhouyanzhao215@mails.ucas.ac.cn}
\and
Gen Li\\
{\tt\small ligen.li@pku.edu.cn}
\and
Nan Duan\\
{\tt\small nanduan@microsoft.com}
\and
Duyu Tang\\
{\tt\small dutang@microsoft.com}
\and
Xiaojie Wang\\
{\tt\small xjwang@bupt.edu.cn}
}

\maketitle

\begin{abstract}
This paper presents a strong baseline for real-world visual reasoning (GQA), which achieves 60.93\% in GQA 2019 challenge and won the sixth place. GQA is a large dataset with 22M questions involving spatial understanding and multi-step inference. To help further research in this area, we identified three crucial parts that improve the performance, namely: multi-source features, fine-grained encoder, and score-weighted ensemble. We provide a series of analysis on their impact on performance.
\end{abstract}

\section{Introduction}

Visual Question Answering (VQA) aims to select an answer given an image and a related question~\cite{Malinowski_multiworldapproachquestion_2014}. It requires both scene understanding in computer vision and semantic understanding in natural language processing. However, previous VQA datasets~\cite{Ren_Exploringmodelsdata_2015,Antol_VqaVisualquestion_2015,Goyal_MakingVQAmatter_2017} are often severely biased and lack semantic compositionality, which makes it hard to diagnose model performance. To handle this, GQA dataset~\cite{hudsonGQANewDataset2019} is proposed. It is more balanced and contains 22M questions that require a diverse set of reasoning skills to answer. 

\begin{figure*}[htbp]
	\centering
	\includegraphics[width=0.88\textwidth]{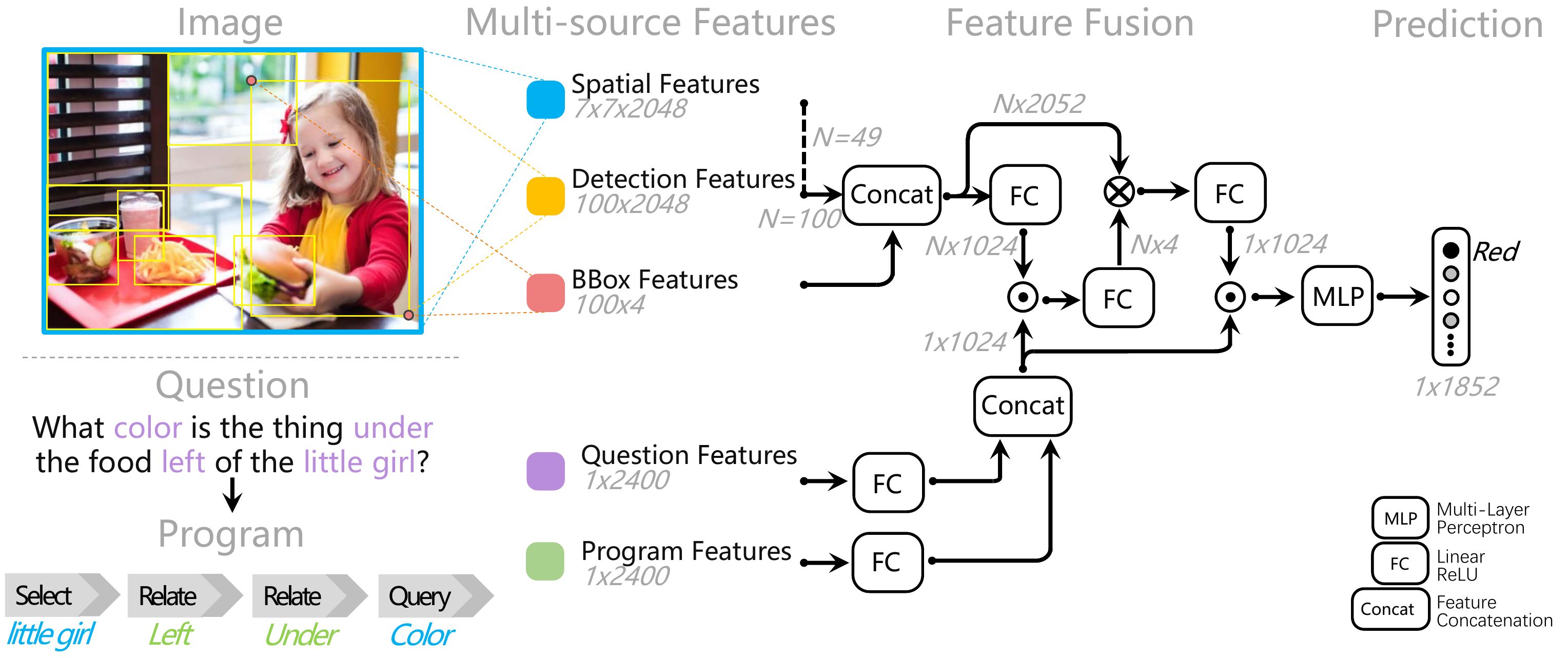}
	\caption{The overall structure of the proposed model for solving the GQA task. We use multi-source features, fine-grained encoder, and weighted ensemble to improve the performance.}
	\label{fig:f1}
\end{figure*}

In the last few years, some novel and interesting approaches have been published to solve the VQA task. For example, the relation-based methods\cite{Santoro_simpleneuralnetwork_2017,wuChainReasoningVisual2018}, attention-based methods~\cite{Kim_Hadamardproductlowrank_2017,wuObjectDifferenceAttentionSimple2018}, and module-based methods~\cite{johnsonInferringExecutingPrograms2017,Andreas_Learningcomposeneural_2016}. In this work, we use a relatively simple architecture as our baseline with three parts, namely: multi-source features, fine-grained encoder, and weighted ensemble. Each part significantly improves performance. Fig.~\ref{fig:f1} provides an overview of our architecture. Firstly, we consider using multi-source features. For images, we use three kinds of features: spatial features, detection features, and bounding box features. For questions, we use both question strings and programs. Secondly, we use Bayesian GRU to encode the question instead of traditional GRU encoder. Thirdly, we use score-weighted ensemble to combine several models. We perform detailed ablation studies of each component which shed lights on developing a strong baseline on real-world visual question answering task. Our model won the sixth place in the 2019 GQA Challenge.

\section{Multi-source features}
To extract features of various aspects of the image, we use three types of features: detection features, spatial features, and bounding box features. The three features are introduced one-by-one below.

\subsection{Multi-source image features}

\subsubsection{Incorporating better detection features}

Better detection features often help better understanding of images. Here we try three detection features: objects features, bottom-up-attention features, and Pythia features. All of them have the same size of 100*2048. The official GQA dataset~\cite{hudsonGQANewDataset2019} provides object features. Bottom-up-attention features are proposed by~\cite{andersonBottomupTopdownAttention2018} who won the first place in the 2017 VQA Challenge. Pythia features are provided by~\cite{singhVQAModelsThat2019}, who is the VQA 2018 challenge winner.

\begin{table}[h]
	\centering
	\begin{tabular}{ccc}
	\specialrule{0.12em}{1pt}{1pt}
		Models&Validation\\
		\hline
        Baseline with object features&62.64\\	
		\hline
        Baseline with bottom-up-attention features&65.44\\
		\hline
		Baseline with pythia features&\textbf{65.99}\\
	\specialrule{0.12em}{1pt}{1pt}	
	\end{tabular}
	\caption{Study better detection features.}\label{tab:t1}
\end{table}

As we see in Tab~\ref{tab:t1}, Pythia features perform better than bottom-up-attention features, and they have a significant gain than object features for about 3\%.

\subsubsection{Adding spatial features}
Some previous work believes that spatial features and detection features may provide complementary information~\cite{Lu_CoattendingFreeformRegions_2018}. In this work, we feed these two features for the two separate pipelines and finally combine the output. As shown in Tab~\ref{tab:t2}, using both detection and spatial features improves the performance for 1\%.
\begin{table}[h]
	\centering
	\begin{tabular}{ccc}
	\specialrule{0.12em}{1pt}{1pt}
		Models&Validation\\
		\hline
        Baseline with detection features&62.64\\	
		\hline
        Baseline with detection and spatial features&\textbf{63.64}\\
	\specialrule{0.12em}{1pt}{1pt}	
	\end{tabular}
	\caption{Study adding spatial features.}\label{tab:t2}
\end{table}

\subsubsection{Adding bounding box features}
One of the drawbacks of the attention method is that it ignores the position information of the objects. Here, We normalize the coordinates of the center point of the bounding box as positional information. Similarly, we normalize the length and width of the bounding box as size information. As shown in Tab~\ref{tab:t3}, using positional information improves performance by 1.13\%. However, the performance gains from size information are not very significant.

\begin{table}[h]
	\centering
	\begin{tabular}{ccc}
	\specialrule{0.12em}{1pt}{1pt}
		Models&Validation\\
		\hline
        Baseline&62.64\\	
		\hline
        Baseline with position features&63.71\\
		\hline
		Baseline with position and size features&\textbf{63.82}\\
	\specialrule{0.12em}{1pt}{1pt}
	\end{tabular}
	\caption{Study adding bound box features.}\label{tab:t3}
\end{table}

\subsection{Multi-source question features}
We use two different ways to represent the semantic meaning of the question. First, we directly use GRU to encode the feature representation $F_{q}$ of the question by feeding in the embedding of words. Second, we develop a VQA domain specific grammar and train a structure prediction model to translate each natural language question into a semantic program which implies the necessary steps for deriving the answer. We then use GRU to encode the program as a feature $F_{p}$. At last, we concatenate $F_{q}$ and $F_{p}$ to form the final representation of the question. The adding of program representation yields about 1\% accuracy improvement on the GQA test split.

\section{Fine-grained encoder}
Here, we use Bayesian GRU to encode the question instead of traditional GRU encoder. Tab.~\ref{tab:t4} shows the performance of different question encoder. As we can see, using baysian GRU improves the performance of about 0.93\%.

\begin{table}[h]
	\centering
	\begin{tabular}{ccc}
	\specialrule{0.12em}{1pt}{1pt}
		Models&Validation\\
		\hline
        Baseline with gru&58.63\\	
		\hline
        Baseline with elmo&59.05\\
		\hline
		Baseline with baysian gru&\textbf{59.56}\\
	\specialrule{0.12em}{1pt}{1pt}		
	\end{tabular}
	\caption{Study Fine-grained encoder.}\label{tab:t4}
\end{table}

\section{Weighted ensemble}
Followed by the early works like \cite{andersonBottomupTopdownAttention2018,singhVQAModelsThat2019}, we use the common practice of ensembling several models to obtain better performance. We choose the best ones of all settings above and try different weights when summing the prediction scores. Both the average ensembling and the best-weighted ensembling results on test-dev splits and the test splits show in Tab.~\ref{tab:t5}. This weighted ensembling strategy improves performance by 1.30\% than the best single model.
\begin{table}[h]
	\centering
	\begin{tabular}{ccc}
	\specialrule{0.12em}{1pt}{1pt}
		Models&Test-dev&Test\\
		\hline
		Average Ensemble&80.08&60.73\\
		\hline
		Weighted Ensemble&81.39&\textbf{60.93}\\
	\specialrule{0.12em}{1pt}{1pt}
	\end{tabular}
	\caption{Study different ensemble strategies.}\label{tab:t5}
\end{table}

{\small
\bibliographystyle{ieee_fullname}
\bibliography{egbib}
}

\end{document}